\documentclass{sigkddExp}
\begin{document}
%
% --- Author Metadata here ---
% -- Can be completely blank or contain 'commented' information like this...
%\conferenceinfo{WOODSTOCK}{'97 El Paso, Texas USA} % If you happen to know the conference location etc.
%\CopyrightYear{2001} % Allows a non-default  copyright year  to be 'entered' - IF NEED BE.
%\crdata{0-12345-67-8/90/01}  % Allows non-default copyright data to be 'entered' - IF NEED BE.
% --- End of author Metadata ---

\title{A Survey on Dialogue Systems:\\ Recent Advances and New Frontiers}
%\subtitle{[Extended Abstract]
% You need the command \numberofauthors to handle the "boxing"
% and alignment of the authors under the title, and to add
% a section for authors number 4 through n.
%
% Up to the first three authors are aligned under the title;
% use the \alignauthor commands below to handle those names
% and affiliations. Add names, affiliations, addresses for
% additional authors as the argument to \additionalauthors;
% these will be set for you without further effort on your
% part as the last section in the body of your article BEFORE
% References or any Appendices.

\numberofauthors{1}
%
% You can go ahead and credit authors number 4+ here;
% their names will appear in a section called
% "Additional Authors" just before the Appendices
% (if there are any) or Bibliography (if there
% aren't)

% Put no more than the first THREE authors in the \author command
%%You are free to format the authors in alternate ways if you have more
%%than three authors.

\author{
%
% The command \alignauthor (no curly braces needed) should
% precede each author name, affiliation/snail-mail address and
% e-mail address. Additionally, tag each line of
% affiliation/address with \affaddr, and tag the
%% e-mail address with \email.
%\alignauthor Ben Trovato \\
%       \affaddr{Institute for Clarity in Documentation}\\
%       \affaddr{1932 Wallamaloo Lane}\\
%       \affaddr{Wallamaloo, New Zealand}\\
%       \email{trovato@corporation.com}
\alignauthor Hongshen Chen$^\dagger$, Xiaorui Liu$^\ddagger$, Dawei Yin$^\dagger$, and Jiliang Tang$^\ddagger$\\
       $^\dagger$\affaddr{Data Science Lab, JD.com}\\
		$^\ddagger$\affaddr{Data Science and Engineering Lab, Michigan State University} \\
       \email{chenhongshen@jd.com, yindawei@acm.org,\{xiaorui,tangjili\}@msu.edu}
}
%\additionalauthors{Additional authors: John Smith (The Th{\o}rvald Group,
%email: {\texttt{jsmith@affiliation.org}}) and Julius P.~Kumquat
%(The Kumquat Consortium, email: {\texttt{jpkumquat@consortium.net}}).}
\date{30 July 1999}
\maketitle
\begin{abstract}
%This paper provides a sample of a LaTeX document for final submission
%to Sigkdd Explorations, the official newsletter of ACM Sigkdd. This is
%a modified version of the ACM Proceedings sample file.

%The developers have tried to include every imaginable sort
%of ``bells and whistles", such as a subtitle, footnotes on
%title, subtitle and authors, as well as in the text, and
%every optional component (e.g. Acknowledgements, Additional
%Authors, Appendices), not to mention examples of
%equations, theorems, tables and figures.

%To make best use of this sample document, run it through \LaTeX\
%and BibTeX, and compare this source code with the printed
%output produced by the dvi file.
Dialogue systems have attracted more and more attention. Recent advances on dialogue systems are overwhelmingly contributed by deep learning techniques, which have been employed to enhance a wide range of big data applications such as computer vision, natural language processing, and recommender systems. For dialogue systems, deep learning can leverage a massive amount of data to learn meaningful feature representations and response generation strategies, while requiring a minimum amount of hand-crafting. In this article, we give an overview to these recent advances on dialogue systems from various perspectives and discuss some possible research directions. In particular, we generally divide existing dialogue systems into task-oriented and non-task-oriented models, then detail how deep learning techniques help them with representative algorithms and finally discuss some appealing research directions that can bring the dialogue system research into a new frontier. 
\end{abstract}

\section{Introduction}
To have a virtual assistant or a chat companion system with adequate intelligence has seemed illusive, and might only exist in Sci-Fi movies for a long time. Recently, human-computer conversation has attracted increasing attention due to its promising potentials and alluring commercial values. With the development of big data and deep learning techniques, the goal of creating an automatic human-computer conversation system, as our personal assistant or chat companion, is no longer an illusion. On the one hand, nowadays we can easily access ``big data" for conversations on the Web and we might be able to learn how to respond and what to respond given (almost) any inputs, which greatly allows us to build data-driven, open-domain conversation systems between humans and computers. On the other hand, deep learning techniques have been proven to be effective in capturing complex patterns in big data and have powered numerous research fields such as computer vision, natural language processing and recommender systems. Hence, a large body of literature has emerged to leverage a massive amount of data via deep learning to advance dialogue systems in many perspectives. 

\begin{figure}[!t]
\includegraphics[width=1.0\linewidth]{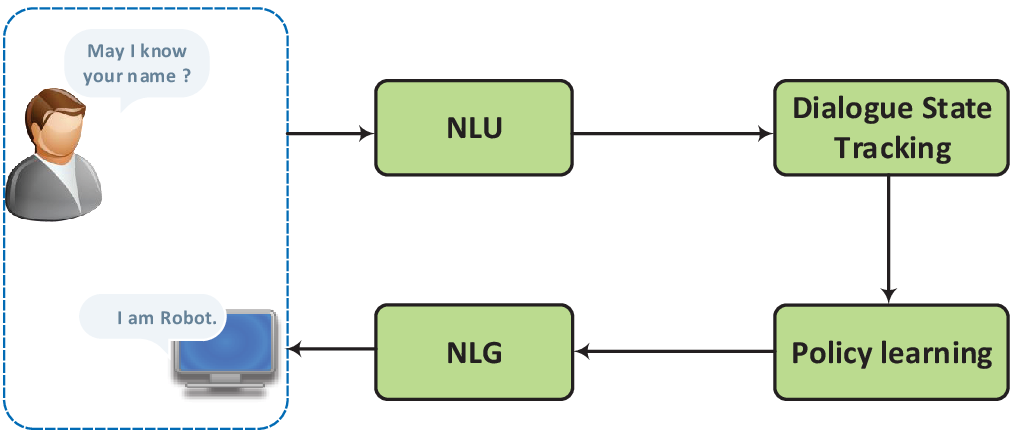}
\caption{Traditional Pipeline for Task-oriented Systems.}
\label{fig:tradl}
\end{figure}
  
According to the applications, dialogue systems can be roughly categorized into two groups -- (1) task-oriented systems and (2) non-task-oriented systems (also known as chat bots). Task-oriented systems aim to assist the user to complete certain tasks (e.g. finding products, and booking accommodations and restaurants). The widely applied approaches to task-oriented systems are to treat the dialogue response as a pipeline as shown in Figure~\ref{fig:tradl}. The systems first understand the message given by human, represent it as a internal state, then take some actions according to the policy with respect to the dialogue state, and finally the action is transformed to its surface form as a natural language. Though language understanding is processed by statistical models, most deployed dialogue systems still use manual features or hand crafted rules for the state and action space representations, intent detection, and slot filling. This not only makes it expensive and time-consuming to deploy a real dialogue system, but also limits its usage to other domains. Recently, many algorithms based on deep learning have been developed to alleviate those problems by learning feature representations in a high dimensional distributed fashion and achieve remarkable improvements in these aspects. In addition, there are attempts to build end-to-end task-oriented dialogue systems, which can expand the state space representation in the traditional pipeline systems and help generalize dialogues outside the annotated task-specific corpora.  

Non-task-oriented systems interact with human to provide reasonable responses and entertainment. Typically, they focus on conversing with human on open domains. Though non-task-oriented systems seem to perform chit-chat, it dominates in many real word applications. As revealed in \cite{zhao2017aaai}, nearly 80\% utterances are chi-chat messages in the online shopping scenario and handling those queries is closely related to user experiences. In general, two major approaches have been developed for non-task-oriented systems -- (1) generative methods such as sequence-to-sequence models, which generate proper responses during the conversation; and (2) retrieval-based methods, which learn to select responses from the current conversation from a repository. 

The recent development of big data and deep learning techniques has greatly advanced both task-oriented and non-oriented dialogue systems, which has encouraged a huge amount of deep learning based researches in dialogue systems. In this article, we aim to (1) give an overview about dialogue systems especially recent advances from deep learning; and (2) discuss possible research directions. The remaining of the article is organized as follows. We review task-oriented dialogue systems including pipeline and end-to-end methods in Section \ref{sec:taskori}. In Section \ref{sec:nontaskori}, we first introduce neural generative methods including popular models and hot research topics; and then detail the retrieval-based methods. In Section \ref{sec:dis}, we conclude the work with discussions on some research directions.

\section{Task-Oriented Dialogue Systems}
\label{sec:taskori}

Task-oriented dialogue systems have been an important branch of spoken dialogue systems. In this section, we will review pipeline and end-to-end methods for task-oriented dialogue systems. 

\subsection{Pipeline Methods}
The typical structure of a pipeline based task-oriented dialogue system is demonstrated in Figure \ref{fig:tradl}. It consists of four key components: 
\begin{itemize} 
\item Language understanding. It is known as natural language understanding (NLU), which parses the user utterance into predefined semantic slots.
\item Dialogue state tracker. It manages the input of each turn along with the dialogue history and outputs the current dialogue state.
\item Dialogue policy learning. It learns the next action based on current dialogue state.
\item Natural language generation (NLG). It maps the selected action to its surface and generates the response.
\end{itemize}

In the following subsections, we will give more details about each component with the state-of-the art algorithms. 

\begin{table}[!t]
\scriptsize
{
  \centering
  \begin{tabular}{|c|c|c|c|c|c|c|}
  \hline
  \textbf{Sentence}     &  show &restaurant&at    &  New   & York &tomorrow\\
  \hline
  \textbf{Slots}        &    O  &  O       &   O  &  B-desti &I-desti &B-date \\
  \hline
  %\textbf{Named Entity} &    O  &  O       &   O  &  B-loc   &I-loc   &B-date \\
  %\hline
  \textbf{Intent}       &    \multicolumn{6}{c|}{Find Restaurant}                   \\
  \hline
  \textbf{Domain}       &    \multicolumn{6}{c|}{Order}                        \\
  \hline
  \end{tabular}
  \caption{An Illustrative Example of Natural Language Representation.}
  \label{tab:exarep}
  }
\end{table}

 \subsubsection{Language Understanding}
Given an utterance, natural language understanding maps it into semantic slots. The slots are pre-defined according to different scenarios. Table \ref{tab:exarep} illustrates an example of natural language representation, where ``New York'' is the location specified as slot values, and the domain and intent are also specified, respectively.
   Typically, there are two types of representations. One is the utterance level category, such as the user's intent and the utterance category. 
   The other is the word-level information extraction such as named entity recognition and slot filling. 
   
   An intent detection is performed to detect the intent of a user. It classifies the utterance into one of the pre-defined intents. Deep learning techniques have been successively applied in intent detection~\cite{deng2012use,tur2012towards,yann2014zero}. In particular, \cite{hashemi2016query} used convolutional neural networks (CNN) to extract query vector representations as features for query classification. The CNN-based classification framework also resembled \cite{huang2013learning} and \cite{shen2014learning}. Similar approaches are also utilized in category or domain classification. 
   
   Slot filling is another challenging problem for spoken language understanding. Unlike intent detection, slot filling is usually defined as a sequence labeling problem, where words in the sentence are assigned with semantic labels. The input is the sentence consisting of a sequence of words, and the output is a sequence of slot/concept IDs, one for each word. \cite{Deoras2013Deep} and \cite{deng2012use} used deep belief networks (DBNs), and achieved superior results compared to CRF baselines. \cite{Mesnil2013Investigation,Yao2013Recurrent,Sarikaya2011Deep,Yao2014Spoken} applied RNN for slot filling. The semantic representation generated by NLU is further processed by the dialogue management component. A typical dialogue management component includes two stages -- dialogue state tracking and policy learning.
 
\subsubsection{Dialogue State Tracking}
Tracking dialogue states is the core component to ensure a robust manner in dialog systems. It estimates the user’s goal at every turn of the dialogue.
 A dialogue state $H_{t}$ denotes the representation of the dialogue session till time $t$. This classic state structure is commonly called slot filling or semantic frame.
 The traditional methods, which have been widely used in most commercial implementations, often adopt hand-crafted rules to select the most likely result \cite{goddeau1996form}. However, these rule-based systems are prone to frequent errors as the most likely result is not always the desired one~\cite{williams2014web}.

A statistical dialog system maintains a distribution over multiple hypotheses of the true dialog state, facing with noisy conditions and ambiguity \cite{Young2010The}.
In Dialog State Tracking Challenge (DSTC) \cite{Williams2012A,williams2013dialog}, the results are in the form of a probability distribution over each slot for each turn.
 A variety of statistical approaches, including robust sets of hand-crafted rules \cite{wang2013simple}, conditional random fields \cite{lee2013recipe,lee2013structured,ren2013dialog}, maximum entropy models \cite{williams2013multi} and web-style ranking \cite{williams2014web} have emerged in Dialog State Tracking Challenge (DSTC) shared tasks.

Recently, \cite{henderson2013deep} introduced deep learning in belief tracking. It used a sliding window to output a sequence of probability distributions over an arbitrary number of possible values. Though it was trained in one domain, it can be easily transferred to new domains. \cite{mrkvsic2015} developed multi-domain RNN dialog state tracking models. It first used all the data available to train a very general belief tracking model, and then specialized the general model for each domain to learn the domain-specific behavior. \cite{mrkvsic-EtAl:2017:Long} proposed a neural belief tracker (NBT) to detect the slot-value pairs. It took the system dialogue acts preceding the user input, the user utterance itself, and a single candidate slot-value pair which it needs to make a decision about, as input, and then iterated over all candidate slot-value pairs to determine which ones have just been expressed by the user.
 
 \subsubsection{Policy learning}
 Conditioned on the state representation from the state tracker, the policy learning is to generate the next available system action. Either supervised learning or reinforcement learning can be used to optimize policy learning \cite{Cuay2015Strategic}. %{\bf can we add some references here} 
 Typically, a rule-based agent is employed to warm-start the system \cite{zhao2017aaai}. Then, supervised learning is conducted on the actions generated by the rules. In online shopping scenario, if the dialogue state is ``Recommendation", then the ``Recommendation" action is triggered, and the system will retrieve products from the product database. If the state is ``Comparison", then the system will compare target products/brands\cite{zhao2017aaai}. The dialogue policy can be further trained end-to-end with reinforcement learning to lead the system making policies toward the final performance.
 \cite{Cuay2015Strategic} applied deep reinforcement learning on strategic conversation that simultaneously learned the feature representation and dialogue policy, the system outperformed several baselines including random, rule-based, and supervised-based methods.
 
 \subsubsection {Natural Language Generation}
 
 The natural language generation component converts an abstract dialogue action into natural language surface utterances. As noticed in ~\cite{Stent2005Evaluating}, a good generator usually relies on several factors: adequacy, fluency, readability, and variation. 

 Conventional approaches to NLG typically perform sentence planning. It maps input semantic symbols into the intermediary form representing the utterance such as tree-like or template structures, and then converts the intermediate structure into the final response through the surface realization \cite{walker2002training,stent2004trainable}.

 \cite{wen-EtAl:2015:W15-46} and \cite{wen-EtAl:2015:EMNLP} introduced neural network-based (NN) approaches to NLG with a LSTM-based structure similar with RNNLM \cite{mikolov2010recurrent}. The dialogue act type and its slot-value pairs are transformed into a 1-hot control vector and is given as the additional input, which ensures that the generated utterance represents the intended meaning.
 \cite{wen-EtAl:2015:W15-46} used a forward RNN generator together with a CNN reranker, and backward RNN reranker. All the sub-modules are jointly optimized to generate utterances conditioned by the required dialogue act. To address the slot information omitting and duplicating problems in surface realization, \cite{wen-EtAl:2015:EMNLP} used an additional control cell to gate the dialogue act. \cite{Tran2017Semantic} extended this approach by gating the input token vector of LSTM with the dialogue act. It was then extended to the multi-domain setting by multiple adaptation steps \cite{wen-EtAl:2016:N16-1}. \cite{zhou2016context} adopted an encoder-decoder LSTM-based structure to incorporate the question information, semantic slot values, and dialogue act type to generate correct answers. It used the attention mechanism to attend to the key information conditioned on the current decoding state of the decoder.  Encoding the dialogue act type embedding, the neural network-based model is able to generate variant answers in response to different act types.
 \cite{duvsek-jurcicek:2016:P16-2} also presented a natural language generator based on the sequence-to-sequence approach that can be trained to produce natural language strings as well as deep syntax dependency trees from input dialogue acts.
 It was then extended with the preceding user utterance and responses~\cite{duvsek-jurcicek:2016:SIGDIAL}. It enabled the model entraining (adapting) to users’ ways of speaking, which provides contextually appropriate responses.
 
 \subsection{End-to-End Methods}
\label{sec:end2endtaskori}
   Despite a lot of domain-specific handcrafting in traditional task oriented dialogue systems, which are diffcult to adapt to new domains \cite{Bordes2016Learning}, \cite{zhao-eskenazi:2016:SIGDIAL} further noted that, the conventional pipeline of task-oriented dialogue systems has two main limitations. One is the credit assignment problem, where the end user's feedback is hard to be propagated to each upstream module. The second issue is process interdependence. The input of a component is dependent on the output of another component. When adapting one component to new environment or retrained with new data, all the other components need to be adapted accordingly to ensure a global optimization. Slots and features might change accordingly. This process requires significant human efforts.
   
   With the advance of end-to-end neural generative models in recent years, many attempts have been made to construct an end-to-end trainable framework for task-oriented dialogue systems. Note that more details about neural generative models will be discussed when we introduce the non-task-oriented systems. Instead of the traditional pipeline, the end-to-end model uses a single module and interacts with structured external databases.
   \cite{wen-EtAl:2017:EACLlong} and \cite{Bordes2016Learning} introduced a network-based end-to-end trainable task-oriented dialogue system, which treated dialogue system learning as the problem of learning a mapping from dialogue histories to system responses, and applied an encoder-decoder model to train the whole system. 
    However, the system is trained in a supervised fashion -- not only does it require a lot of training data, but it may
also fail to find a good policy robustly due to the lack of exploration of dialogue control in the training data. \cite{zhao-eskenazi:2016:SIGDIAL} first presented an end-to-end reinforcement learning approach to jointly train dialogue state tracking and policy learning in the dialogue management in order to optimize the system actions more robustly. In the conversation, the agent asks the user a series of Yes/No questions to find the correct answer. This approach was shown to be promising when applied to the task-oriented dialogue problem of guessing the famous people users think of. \cite{li2017end} trained the end-to-end system as a task completion neural dialogue system, where its final goal is to complete a task, such as movie-ticket booking.

Task-oriented systems usually need to query outside knowledge base. Previous systems achieved this by issuing a symbolic query to the knowledge base to retrieve entries based on their attributes, where semantic parsing on the input is performed to construct a symbolic query representing the beliefs of the agent about the user goal\cite{wen-EtAl:2017:EACLlong,williams2016end,li2017end}. This approach has two drawbacks: (1) the retrieved results do not carry any information about uncertainty in semantic parsing, and (2) the retrieval operation is non differentiable, and hence the parser and dialog policy are trained separately. This makes online end-to-end learning from user feedback difficult once the system is deployed. \cite{eric2017key} augmented existing recurrent network architectures with a differentiable attention-based key-value retrieval mechanism over the entries of a knowledge base, which is inspired by key-value memory networks\cite{miller-EtAl:2016:EMNLP2016}.
    \cite{dhingra-EtAl:2017:Long1} replaced symbolic queries with an induced ``soft'' posterior distribution over the knowledge base that indicates which entities the user is interested in. Integrating the soft retrieval process with a reinforcement learner.
    \cite{williams-asadi-zweig:2017:Long} combined an RNN with domain-specific knowledge encoded as software and system action templates.

\section{Non-Task-Oriented Dialogue System}
\label{sec:nontaskori}

Unlike task-oriented dialogue systems, which aim to complete specific tasks for user, non-task-oriented dialogue systems (also known as chatbots) focus on conversing with human on open domains \cite{Ritter2011Data}. In general, chat bots are implemented either by generative methods or retrieval-based methods. Generative models are able to generate more proper responses that could have never appeared in the corpus, while retrieval-based models enjoy the advantage of informative and fluent responses\cite{ji2014information}, because they select a proper response for the current conversation from a repository with response selection algorithms.
In the following sections, we will first dive into the neural generative models, one of the most popular research topics in recent years, and discuss their drawbacks and possible improvements. Then, we introduce recent advances of deep learning in retrieval based models.

\subsection{Neural Generative Models}
 Nowadays, a large amount of conversational exchanges is available in social media websites such as Twitter and Reddit, which raise the prospect of building data-driven models. 
 \cite{Ritter2011Data} proposed a generative probabilistic model, which is based on phrase-based Statistical Machine Translation \cite{Zens2002Phrase}, to model conversations on micro-blogging. It viewed the response generation problem as a translation problem, where a post needs to be translated into a response. 
  However, generating responses was found to be considerably more difficult than translating between languages. It is likely due to the wide range of plausible responses and the lack of phrase alignment between the post and the response. The success of applying deep learning in machine translation, namely Neural Machine Translation, spurs the enthusiasm of researches in neural generative dialogue systems.
  
In the following subsections, we first introduce the sequence-to-sequence models, the foundation of neural generative models. Then, we discuss hot research topics in the direction including incorporating dialogue context, improving the response diversity, modeling topics and personalities, leveraging outside knowledge base, the interactive learning and evaluation. 
  
\begin{figure}[!t]
\includegraphics[width=1.0\linewidth]{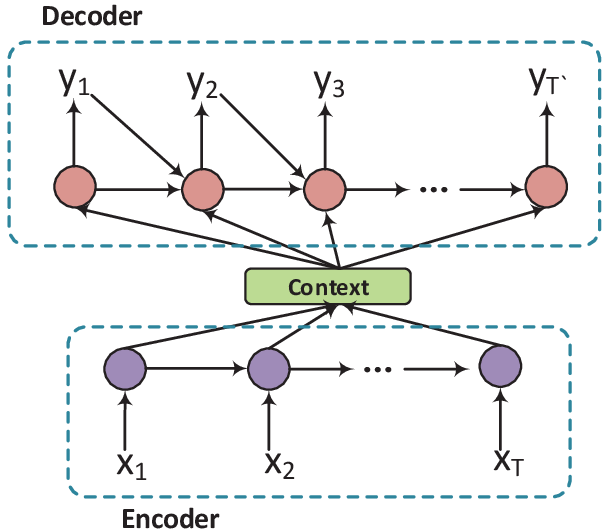}
\caption{An Illustration of the Encoder-Decoder Model.}
\label{fig:encdec}
\end{figure}

\subsubsection{Sequence-to-Sequence Models}
    Given a source sequence (\emph{message}) $X = (x_{1}, x_{2},..., x_{T})$ consisting of $T$ words and a target sequence (\emph{response}) $Y = (y_{1}, y_{2},..., y_{T^{'}} )$ of length $T^{'}$,% {\bf what do ``T" and ``T'" mean}
the model maximizes the generation probability of $Y$ conditioned on $X$: $p(y_{1},..., y_{T^{'}}|x_{1},..., x_{T})$. 
    Specifically, a sequence-to-sequence model (or Seq2Seq) is in an encoder-decoder structure. Figure \ref{fig:encdec} is a general illustration of such structure. The encoder reads $X$ word by word and represents it as a context vector $c$ through a recurrent neural network (RNN),and then the decoder estimates the generation probability of $Y$ with $c$ as the input. The encoder RNN calculates the context vector $\mathbf{c}$ by
    \begin{equation}
    \nonumber
    \mathbf{h}_{t}=f(x_{t},\mathbf{h}_{t-1}),
    \end{equation}
where $\mathbf{h}_{t}$ is the hidden state at time step $t$, $f$ is a non-linear function such as long-short term memory unit (LSTM) \cite{Graves1997Long} and gated recurrent unit (GRU) \cite{cho-EtAl:2014:EMNLP2014}, and $c$ is the hidden state corresponding to the last word $\mathbf{h}_{T}$.
	The decoder is a standard RNN language model with an additional conditional context vector $\mathbf{c}$. 
    The probability distribution $\mathbf{p}_{t}$ of candidate words at every time $t$ is calculated as
    \begin{flalign}
    \nonumber
    \mathbf{s}_{t} &= f(y_{t-1},\mathbf{s}_{t-1},\mathbf{c}),\\
    \nonumber
    \mathbf{p}_{t} &= softmax(\mathbf{s}_{t},y_{t-1}),
    \end{flalign}
 where $\mathbf{s}_{t}$ is the hidden state of the decoder RNN at time $t$ and $y_{t−1}$ is the word at time $t − 1$ in the response sequence.
    The objective function of Seq2Seq is defined as:
    \begin{equation}
    \nonumber
      p((y_{1},..., y_{T^{'}} |x_{1},..., x_{T})=p(y_{1}|\mathbf{c})\prod_{t=2}^{T^{'}}p(y_{t}|\mathbf{c},y_{1},...,y_{t-1}).
    \end{equation}
   
   \cite{bahdanau2014neural} then improved the performance by the attention mechanism, where each word in $Y$ is conditioned on different context vector $\mathbf{c}$, with the observation that each word in $Y$ may relate to different parts in $x$.
In particular, $y_{i}$ corresponds to a context vector $\mathbf{c}_{i}$, and $\mathbf{c}_{i}$ is a weighted average of the encoder hidden states $\mathbf{h}_{1},...,\mathbf{h}_{T}$:
    \begin{equation}
    \nonumber
    \mathbf{c}_{i}= \Sigma_{j=1}^{T}\alpha_{ij}\mathbf{h}_{j},
    \end{equation}
    where $\alpha_{i,j}$ is computed by:
    \begin{flalign}
    \nonumber
    \alpha &= \frac{exp(e_{ij})}{\Sigma_{k=1}^{T}exp(e_{ik})},\\
    \nonumber
    e_{ij} &= g (\mathbf{s}_{t-1},\mathbf{h}_{j}),
    \end{flalign}
    where $g$ is a multilayer perceptron. %{\bf what do you mean ``single-layer, multilayer..''}

  \cite{shang-lu-li:2015:ACL-IJCNLP} applied the recurrent neural network encoder-decoder framework \cite{cho-EtAl:2014:EMNLP2014} to generate responses on Twitter-style micro-blogging websites, while \cite{vinyals2015neural} utilized a similar model described in \cite{luong-EtAl:2015:ACL-IJCNLP}. 
   In general, these models utilize neural networks to represent dialogue histories and to generate appropriate responses. Such models are able to leverage a large amount of data in order to learn meaningful natural language representations and generation strategies, while requiring a minimum amount of domain knowledge and hand-crafting. 
  
\subsubsection{Dialogue Context}
  The ability to take into account previous utterances is key to building dialog systems that can keep conversations active and engaging. 
%  Figure X illustrates a typical Twitter dialog where the contextual information is crucial: the phrase “good luck” is plainly motivated by the reference to “your game” in the first utterance \cite{sordoni-EtAl:2015:NAACL-HLT}. 
   \cite{sordoni-EtAl:2015:NAACL-HLT} addressed the challenge of the context sensitive response generation by representing the whole dialogue history (including the current message ) with continuous representations or embeddings of words and phrases. The response is generated as RNN language model \cite{mikolov2010recurrent}, the same as the decoder in \cite{cho-EtAl:2014:EMNLP2014}. %\cite{mikolov2010recurrent}, and the representation of context and message are computed through a bag-of-words representation.
   \cite{AAAI1611957} used hierarchical models, first capturing the meaning of individual utterances and then integrating them as discourses. \cite{Xing2017Hierarchical} extended the hierarchical structure with the attention mechanism \cite{bahdanau2014neural} to attend to important parts within and among utterances with word level attention and utterance level attention, respectively. \cite{Tian2017How} conducted a systematic comparison among existing methods (including non-hierarchical and hierarchical models) and proposed a variant that weights the context with respect to context-query relevance. It found that (1) hierarchical RNNs generally outperform non-hierarchical ones, and (2) with context information, neural networks tend to generate longer, more meaningful and diverse replies.
 
\subsubsection{Response Diversity}
    A challenging problem in current sequence-to-sequence dialogue systems is that they tend to generate trivial or non-committal, universally relevant responses with little meaning, which are often involving high frequency phrases along the lines of \emph{I don’t know} or \emph{I’m OK} \cite{sordoni-EtAl:2015:NAACL-HLT,vinyals2015neural,AAAI1611957}. 
    
    This behavior can be ascribed to the relative high frequency of generic responses like \emph{I don’t know} in conversational datasets, in contrast with the relative sparsity of more informative alternative responses. 
   One promising approach to alleviate such challenge is to find a better objective function. 
   \cite{li-EtAl:2016:N16-11} pointed out that neural models assign high probability to ``safe” responses when optimizing the likelihood of outputs given inputs. They used a Maximum Mutual Information (MMI), which was first introduced in speech recognition \cite{bahl1986maximum,brown1987acoustic}, as an optimization objective. It measured the mutual dependence between inputs and outputs, where it took into consideration the inverse dependency of responses on messages.
   \cite{Yao2016An} incorporated inverse document frequency (IDF) \cite{Salton1988Term} into the training process to measure the response diversity.
     
   Some researches realized that the decoding process is another source of redundant candidate responses.
   \cite{vijayakumar2016diverse},\cite{Shao2017Generating} and \cite{Li2016A} recognized that the beam-search,  an approximate inference algorithm to decode output sequences for neural sequence models, lacks diversity when generating candidates in the beam. 
   \cite{vijayakumar2016diverse} augmented the beam-search objective with a dissimilarity term that measured the diversity between candidate sequences.
   \cite{Shao2017Generating} introduced a stochastic beam-search procedure, while \cite{Li2016A} added an additional term for beam search scoring to penalize the siblings--expansions of the same parent node in the search. 
   \cite{li-EtAl:2016:N16-11,sordoni-EtAl:2015:NAACL-HLT,Shao2017Generating} further performed a re-ranking step with global features to avoid generating dull or generic responses. 
   \cite{mou-EtAl:2016:COLING} conjectured that not only the problem lies in the objective of decoding and response frequency, but also the message itself may lack sufficient information for the replay. It proposed to use pointwise mutual information (PMI) to predict a noun as a keyword, reflecting the main gist of the reply, and then generates a reply containing the given keyword.  

   Another series of works have focused on generating more diverse outputs by introducing a stochastic latent variable. They demonstrated that natural dialogue is not deterministic -- replies for a same message may vary from person to person. However, current response is sampled from a deterministic encoder-decoder model. By incorporating a latent variable, these models have the advantage that, at the generation time, they can sample a response from the distribution by first sampling an assignment of the latent variables, and then decoding deterministically. 
   \cite{cao-clark:2017:EACLshort} presented a latent variable model for one-shot dialogue response. The model contained a stochastic component $z$ in the decoder $P(Y |z, X)$, where $z$ is computed following the variational auto-encoder framework\cite{Kingma2013Auto,Kingma2014Semi,Sohn2015Learning}. 
   \cite{hvred} introduced latent variables to the hierarchical dialogue modeling framework \cite{AAAI1611957}. The latent variable is designed to make high-level decisions like topic or sentiment. 
   \cite{shen-EtAl:2017:Short} conditioned the latent variable on explicit attributes to make the latent variable more interpretable. These attributes can be either manually assigned or automatically detected such topics, and personality.

\subsubsection{Topic and Personality}
 Learning the inherent attributes of dialogues explicitly is another way to improve the diversity of dialogues and ensures the consistency. Among different attributes, topic and personality are widely explored. 
 
    \cite{AAAI1714563} noticed that people often associate their dialogues with topically related concepts and create their responses according to these concepts. They used Twitter LDA model to get the topic of the input, fed topic information and input representation into a joint attention module and generated a topic-related response. A small improvement in decoder had achieved a better result in \cite{Xing2016Topic}.
    \cite{Choudhary2017Domain} made a more thorough generalization of the problem. They classified each utterance in the dialogue into one domain, and generated the domain and content of next utterance accordingly.
    
       \cite{AAAI1714571} jointly modeled the high-level coarse tokens sequence and the dialogue generation explicitly, where the coarse tokens sequence aims to exploit high-level semantics. They exploited nouns and activity-entity for the coarse sequence representation.
       
    \cite{Zhou2017Emotional} added emotion embedding into a generative model and achieved good performance in perplexity. 
    \cite{asghar2017affective} enhanced the model of producing emotionally rich responses from three aspects: incorporating cognitive engineered affective word embeddings, augmenting the loss objective with an affect-constrained objective function, and injecting affective dissimilarity in diverse beam-search inference procedure\cite{vijayakumar2016diverse}.
    \cite{Qian2017Assigning} gave the system an identity with profile so that the system can answer personalized question consistently. 
    \cite{li-EtAl:2016:P16-13} further took the information of addressee into consideration to create a more realistic chatbot.

    Since the training data comes from different speakers with inconsistency, \cite{Zhang2017Neural} proposed a two-phase training approach which initialized the model using large scale data and then fine-tuned the model to generate personalized response. 
    \cite{Mo2016Personalizing} used transfer reinforcement learning to eliminate inconsistencies.

\subsubsection{Outside Knowledge Base}
   An important distinction between human conversation and dialogue system is whether it is combined with reality. 
   Incorporating an outside Knowledge Base (KB) is a promising approach to bridge the gap of background knowledge between a dialogue system and human. %This leads to many works, which introduce background knowledge into dialogue systems.
   
   	Memory network is a classic method dealing with question answering tasks with knowledge base. Thus, it is quite straightforward to apply it in dialogue generation. 
	\cite{Ghazvininejad2017A} made attempts on top of this and has achieved good performance in open-domain conversations.
	\cite{vougiouklis-hare-simperl:2016:COLING} also worked on open-domain conversations with background knowledge by coupling CNN embedding and RNN embedding into multimodal space and made progress in perplexity. 
	A similar task is to generate an answer for a question according to outside knowledge. Unlike the general method of tuple retrieval in knowledge base, \cite{Yin:2016:NGQ:3060832.3061037} used words from knowledge base together with common words in generation process. Empirical studies demonstrated that the proposed model was capable of generating natural and right answers to the questions by referring to the facts in the knowledge base.

\subsubsection{Interactive Dialogue learning}
    Learning through interaction is one of the ultimate goals of dialogue systems. 
    \cite{li-EtAl:2016:EMNLP20162} simulated dialogues between two virtual agents. They defined simple heuristic approximations to rewards that characterize good conversations: good conversations are forward-looking \cite{allwood1992semantics} or interactive (a turn suggests a following turn), informative, and coherent. 
    The parameters of an encoder-decoder RNN defined a policy over an infinite action space consisting of all possible utterances. The agent learned a policy by optimizing the long-term developer-defined reward from ongoing dialogue simulations using policy gradient methods \cite{williams1992simple}, rather than the MLE objective defined in the standard SEQ2SEQ models. \cite{li2016dialogue} further attempted to improve the bot's ability to learn from interaction. By using policy learning and forward prediction on both textual and numerical feedback, the model can improve itself by interacting with human in a (semi-)online way. Instead of using hand-crafted reward functions for online reinforcement learning, \cite{asghar2017deep} performed online human in-the-loop active learning by repeatedly letting human select one of the $K$ responses, generated by an offline supervised pretrained dialogue agent, as the 'best' response, and then respond to the selected response. The network is also trained with the joint corss-entropy loss function.
    
    As most human respondents may ask for clarification or hints when not confident about the answer, it is natural to make the bot owning such a capability. \cite{Li2017Learning} defined three situations where the bot has problems in answering a question. Compared the experimental results of not using a asking-question way, this method made great improvement in some scenarios. \cite{lewis-EtAl:2017:EMNLP2017} explored the task on negotiation dialogues. As conventional sequence-to-sequence models simulate human dialogues but fail to optimize a specific goal, this work took a goal-oriented training and decoding approach and demonstrated a worthwhile perspective.
 
\subsubsection{Evaluation}
     Evaluating the quality of the generated response is an important aspect of dialogue response generation
systems\cite{liu-EtAl:2016:EMNLP20163}. Task-oriented dialogue system can be evaluated based on human-generated supervised signals, such as a task completion test or a user satisfaction score\cite{walker1997paradise,moller2006memo,kamm1995user}, however,  automatically evaluating the quality of generated responses for non-task-oriented dialogue systems remains an open question due to the high response diversity \cite{artstein2009semi}.
	Despite the fact that word overlap metrics such as BLEU, METEOR, and ROUGE have been widely used to evaluate the generated responses, \cite{liu-EtAl:2016:EMNLP20163} found that those metrics, as well as word embedding metrics derived from word embedding models such as Word2Vec\cite{mikolov2013distributed} have either weak or no correlation with human judgements, although word embedding metrics are able to significantly distinguish between baselines and state-of-the-art models across multiple datasets. 
    \cite{su2015learning} proposed to use two neural network models to evaluate a sequence of turn-level features to rate the success of a dialogue.
    \cite{lowe-EtAl:2017:Long} encoded the context, the true response and the candidate response into vector representations using RNN, and then computed a score using a dot-product between the vectors in a linearly transformed space. 
    \cite{tao2017ruber} combined referenced and unreferenced metrics, where the former measured the similarity between reply and the groundtruth through word embedding, and the latter scored the correlation between reply and query trained with a max-margin objective function, where the negative reply is randomly sampled.
    
	One promising approach comes from the idea of Turing test\cite{turing1950computing}-- employing an evaluator to distinguish machine-generated texts from human-generated ones. \cite{kannan2017adversarial} and \cite{bruni2017adversarial} explored adversarial evaluation model \cite{bowman2015generating} which assigns a score based on how easy it is to distinguish the generated responses from human responses, while \cite{li-EtAl:2017:EMNLP20175} directly applied adversarial learning\cite{goodfellow2014generative,denton2015deep} into dialogue generation.
 
\subsection{Retrieval-based Methods}
Retrieval-based methods choose a response from candidate responses. The key to retrieval-based methods is message-response matching. Matching algorithms have to overcome semantic gaps between messages and responses~\cite{hu2014convolutional}.
 
\subsubsection{Single-turn Response Matching}
    Early studies of retrieval-based chatbots mainly focus on response selection for single-turn conversation\cite{wang-EtAl:2013:EMNLP3}, where only the message is used to select a proper response.
    Typically, the context and the candidate response are encoded as a vector respectively, and then the matching score is computed based on those two vectors.
    Suppose $\mathbf{x}$ is the vector representation of a message and $\mathbf{y}$ corresponds to the response vector representation, the matching function between $\mathbf{x}$ and $\mathbf{y}$ can be as simply as bilinear matching:
	\begin{equation}
	\nonumber
    match(\mathbf{x},\mathbf{y})=\mathbf{x}^ \mathrm{T} \mathbf{A} \mathbf{y},
	\end{equation}
    where $\mathbf{A}$ is a pre-determined matrix, or more complicated ones.
    \cite{Lu2013A} proposed a DNN-based matching model for short texts response selection and combined the localness and hierarchy intrinsic in the structure. 
    \cite{hu2014convolutional} improved the model by utilizing a deep convolutional neural network architecture to learn the representation of message and response, or directly learn the interacted representation of two sentences, followed by a multi-layer perceptron to compute the matching score. 
    \cite{wang2015ijcai} extracted dependency tree matching patterns and used them as sparse one-hot inputs of a deep feed-forward neural network for context-response matching.
    \cite{Wu2016Topic} incorporated the topic vector generated by Twitter LDA model \cite{Zhao2011Comparing} into the CNN based structure to boost responses with rich content.
  
\subsubsection{Multi-turn Response Matching}
    In recent years, multi-turn retrieval-based conversation draws more and more attention. In multi-turn response selection, current message and previous utterances are taken as input. The model selects a response that is natural and relevant to the whole context.  
    It is important to identify important information in previous utterances and properly model the utterances relationships to ensure conversation consistency. \cite{lowe-EtAl:2015:W15-46} encoded the context (a concatenation of all previous utterances and current message) and candidate response into a context vector and a response vector through a RNN/LSTM based structure, respectively, and then computed the matching degree score based on those two vectors.
    \cite{Yan:2016} selected the previous utterances in different strategies and combined them with current messages to form a reformulated context. 
    \cite{zhou-EtAl:2016:EMNLP20161} performed context-response matching on not only the general word level context vector but also the utterance level context vector. 
    \cite{wu-EtAl:2017:Long1} further improved the leveraging of utterances relationship and contextual information by matching  a response with each utterance in the context on multiple levels of granularity with a convolutional neural network, and then accumulated the vectors in a chronological order through a recurrent neural network to model relationships among utterances. 
 
\subsection{Hybrid Methods}
    Combing neural generative and retrieval based models can have significant effects on performance.\cite{song2016two} and \cite{qiu2017alime} attempted to combine both methods. Retrieval-based systems often give precise but blunt answers, while generation-based systems tend to give fluent but meaningless responses. In an ensemble model, the retrieved candidate, along with the original message, are fed to an RNN-based response generator. The final response is given by a post-reranker. This approach combined the advantages of retrieval and generation based models, which was appealing in performance.
    \cite{serban2017deep} integrated natural language generation and retrieval models, including template-based models, bag-of-words models, sequence-to-sequence neural network and latent variable neural network models and applied reinforcement learning to crowdsourced data and real-world user interactions to select an appropriate response from the models in its ensemble.

\section{Discussion and Conclusion}
\label{sec:dis}
     Deep learning has become a basic technique in dialogue systems. Researchers investigated on applying neural networks to the different components of a traditional task-oriented dialogue system, including natural language understanding, natural language generation, dialogue state tracking. Recent years, end-to-end frameworks become popular in not only the non-task-oriented chit-chat dialogue systems, but also the task-oriented ones. Deep learning is capable of leveraging large amount of data and is promising to build up a unified intelligent dialogue system. It is blurring the boundaries between the task-oriented dialogue systems and non-task-oriented systems. In particular, the chit-chat dialogues are modeled by the sequence-to-sequence model directly. The task completion models are also moving towards an end-to-end trainable style with reinforcement learning representing the state-action space and combing the whole pipelines. 

It is worth noting that current end-to-end models are still far from perfect. Despite the aforementioned achievements, the problems remain challenging. Next, we discuss some possible research directions: 

\begin{itemize}
 \item Swift Warm-Up. Although end-to-end models have drawn most of the recent research attention, we still need to rely on traditional pipelines in practical dialogue engineering, especially in a new domain warm-up stage. The daily conversation data is quite ``big'', however, the dialogue data for a specific domain is quite limited. In particular, domain specific dialogue data collection and dialogue system construction are laborsome. Neural network based models are better at leveraging large amount of data. We need new way to bridge over the warm-up stage. It is promising that the dialogue agent has the ability to learn by itself from the interactions with human. 
\item Deep Understanding. Current neural network based dialogue systems heavily rely on the huge amount of different types of annotated data, and structured knowledge base and conversation data. They learn to speak by imitating a response again and again, just like an infant, and the responses are still lack of diversity and sometimes are not meaningful. Hence, the dialogue agent should be able to learn more effectively with a deep understanding of the language and the real world. Specifically, it remains much potential if a dialogue agent can learn from human instruction to get rid of repeatedly training. Since a great quantity of knowledge is available on the Internet, a dialogue agent can be smarter if it is capable of utilizing such unstructured knowledge resource to make comprehension. Last but not least, a dialogue agent should be able to make reasonable inference, find something new, share its knowledge across domains, instead of repeating the words like a parrot.
\item Privacy Protection. Widely applied dialogue system serves a large number of people. It is quite necessary to notice the fact that we are using the same dialogue assistant. With the ability of learning through interactions, comprehension and inference, a dialogue assistant can inadvertently and implicitly store some of sensitive information~\cite{papernot2016semi}. Hence, it is important to protect users' privacy while building better dialogue systems.
\end{itemize}

\section*{Acknowledgements}
Xiaorui Liu and Jiliang Tang are supported by the National Science Foundation (NSF) under grant number IIS-1714741 and IIS-1715940.

\bibliographystyle{abbrv}
\bibliography{ref}  
\end{document}